%% file: unsup_oneshot_interspeech2020.tex
\documentclass[a4paper]{article}

\usepackage{INTERSPEECH2020}

\usepackage{graphicx}
\graphicspath{{fig/}}

\usepackage{booktabs}
\usepackage{tabularx}
\usepackage{multirow}
\newcommand{\mytable}{
    \centering
    \renewcommand{\arraystretch}{1.1}
    }

\newcolumntype{C}{>{\centering\arraybackslash}X}
\newcolumntype{L}{>{\raggedright\arraybackslash}X}
\newcolumntype{R}{>{\raggedleft\arraybackslash}X}
\newcolumntype{P}[1]{>{\raggedright\arraybackslash}p{#1}}

\usepackage{amsmath}
\usepackage{amssymb}
\renewcommand{\vec}[1]{\boldsymbol{\mathbf{#1}}}

\usepackage{calc}
\usepackage{microtype}
\usepackage{url}
\usepackage{xcolor}

\newcommand\blfootnote[1]{\begingroup
                          \renewcommand\thefootnote{}\footnote{#1}
                          \addtocounter{footnote}{-1}
                          \endgroup}

\usepackage{cite}
\title{Unsupervised vs.\ transfer learning for multimodal one-shot \\ matching of speech and images}
\name{Leanne Nortje \qquad Herman Kamper}
\address{E\&E Engineering, Stellenbosch University, South Africa}
\email{\small nortjeleanne@gmail.com, kamperh@sun.ac.za}

\usepackage[prependcaption,textsize=scriptsize]{todonotes}
\definecolor{mycolor}{HTML}{FF6600}

\begin{document}

\maketitle

\input{abstract}
\input{Introduction/Introduction}
\input{Multimodal_one-shot_learning_task/Multimodal_one-shot_learning_task}
\input{Feature_extraction_methods_for_multimodal_one-shot_learning/Feature_extraction_methods_for_multimodal_one-shot_learning}
\input{Experiments/Experiments}
\input{Conclusion/Conclusion}

\newpage
\bibliography{library}

\end{document}

%% file: abstract.tex
\begin{abstract}
We consider the task of multimodal one-shot speech-image matching. An agent is shown a picture along with a spoken word describing the object in the picture, e.g.\ \textit{cookie}, \textit{broccoli} and \textit{ice-cream}. After observing \textit{one} paired speech-image example per class, it is shown a new set of unseen pictures, and asked to pick the ``ice-cream''. Previous work attempted to tackle this problem using transfer learning: supervised models are trained on labelled background data not containing any of the one-shot classes. Here we compare transfer learning to unsupervised models trained on unlabelled in-domain data. On a dataset of paired isolated spoken and visual digits, we specifically compare unsupervised autoencoder-like models to supervised classifier and Siamese neural networks. In both unimodal and multimodal few-shot matching experiments, we find that transfer learning outperforms unsupervised training. We also present experiments towards combining the two methodologies, but find that transfer learning still performs best (despite idealised experiments showing the benefits of unsupervised learning).
\end{abstract}
\noindent\textbf{Index Terms}: one-shot learning, multimodal modelling, unsupervised models, transfer learning, word acquisition

%% file: Introduction/Introduction.tex
\graphicspath{{Introduction/Figures/}}

\section{Introduction}
\label{sec:introduction}

Young children are able to learn new objects and words from only a few examples~\cite{biederman_recognition-by-components:_1987,miller_how_1987,gomez_infant_2000,rasanen_joint_2015}.
In contrast, most conventional vision or speech processing systems require large amounts of labelled data.
This has motivated studies into one-shot learning~\cite{li_fei-fei_bayesian_2003,fei-fei_one-shot_2006,lake_one_2011, lake_one-shot_2014,koch_siamese_2015,vinyals_matching_2016,shyam_attentive_2017}: to learn a new concept from one or a few labelled examples. 
One-shot learning
studies have mainly focused on learning new concepts in a single modality.
But recently, \textit{multimodal one-shot learning} has also been considered~\cite{eloff_multimodal_2019}.
Instead of observing an item together with a class label, the model observes 
a pair of items coming from different modalities but representing the same concept. 
As an example, imagine a household robot is shown examples of \textit{milk}, \textit{eggs}, \textit{butter} and a \textit{mug}, each visual instance being paired with a spoken tag.
At test time, the agent is then presented with a spoken query such as ``butter'', and asked to identify the corresponding visual object. 

In~\cite{eloff_multimodal_2019}, this was investigated on a dataset of isolated spoken digits paired with images.
To perform multimodal matching at test-time, separate speech-speech and image-image comparisons were combined: a spoken query is compared to all the {speech} items in a so-called \textit{support set}, the image corresponding to the closest item {in the support set} is determined, and this image is then compared to all the items in the \textit{matching set} to predict the test image best matching the input speech query.
To learn a distance metric within each modality, 
transfer learning was used by training supervised vision and speech models on background training data not containing any of the one-shot test classes.
{As in other unimodal one-shot studies in gesture recognition~\cite{thomason_recognizing_2017,wu_one_2012}, video~\cite{stafylakis_zero-shot_2018} and robotics~\cite{walter_one-shot_2012, finn_one-shot_2017},}
this can be motivated by the observation that humans can call on prior knowledge when 
learning new concepts. 

Except for existing knowledge, it is also conceivable that, before being shown paired examples, an agent such as the household robot would be exposed to a large amount of \textit{unlabelled} speech and visual data from its environment.
Some of these unlabelled examples could 
correspond to the classes of interest. 
Motivated by this observation, we ask 
how unsupervised models trained on unlabelled in-domain data compares to transfer learning from background data for multimodal one-shot matching.

To learn feature representations for within-modality comparisons, 
we specifically consider two unsupervised learning strategies.
An autoencoder (AE) attempts to reproduce its input at its output through a bottleneck feature layer. 
The correspondence autoencoder (CAE) tries to reproduce another instance of the input at its output~\cite{kamper_unsupervised_2015}.
Since we only have unlabelled data, the CAE samples nearest neighbours 
to obtain its output targets. 
We compare these unsupervised models to supervised classifier and Siamese neural networks trained on background data~\cite{eloff_multimodal_2019}.
Each of the models are trained separately on vision and speech data and then used to estimate {within-modality} similarity. 

On the same isolated digit speech-image multimodal one-shot matching task as in~\cite{eloff_multimodal_2019}, 
we show that transfer learning outperforms unsupervised modelling.
We also consider 
approaches for combining transfer and unsupervised learning.
Although this 
yields improvements over a
purely unsupervised model, the best overall performance is still achieved through transfer learning.\footnote{\raggedright We release source code at: {\scriptsize \url{https://github.com/LeanneNortje/multimodal_speech-image_matching}}.}

%% file: Multimodal_one-shot_learning_task/Multimodal_one-shot_learning_task.tex
\graphicspath{{Multimodal_one-shot_learning_task/Figures/}}

\begin{figure*}[htb]
        
        \begin{minipage}[b]{1.0\textwidth}
            \centering
                \includegraphics[width=0.85\textwidth]{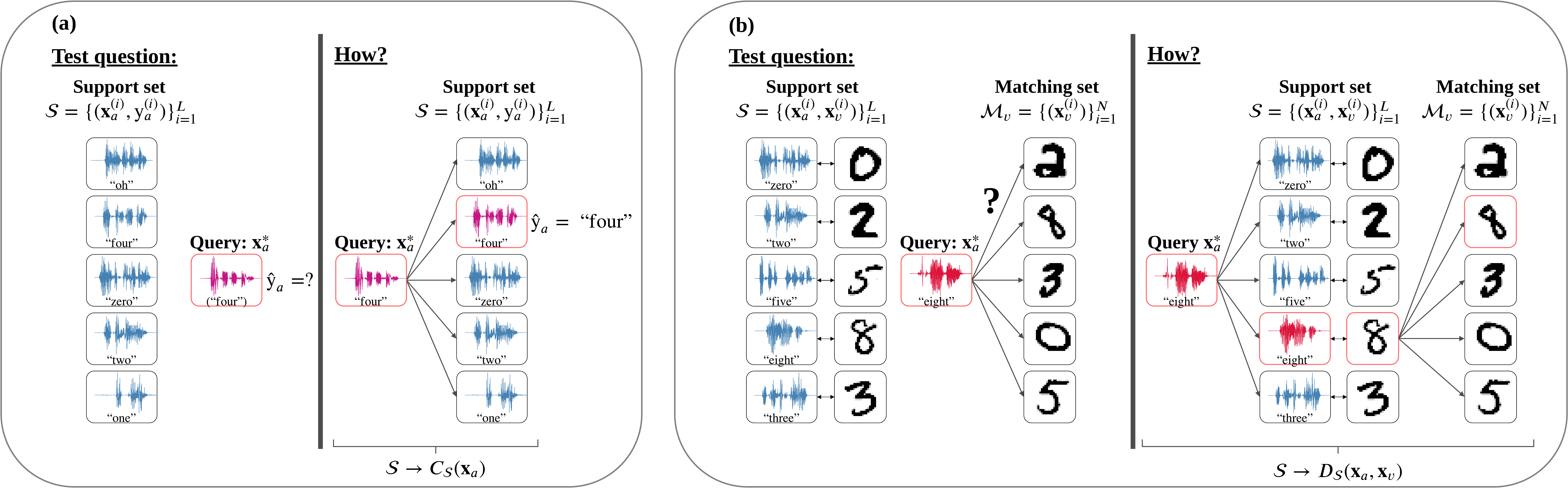}
        \vspace*{-6pt}
        \end{minipage}
        \caption{(a) Unimodal one-shot speech classification 
        and (b) multimodal one-shot speech-image matching. In both cases, the left side illustrates the question shown at test time, and the right side illustrates how the model makes its prediction.}
        \label{fig:unimodal+multimodal}
        
        \vspace*{-12pt}
\end{figure*}

\section{Multimodal one-shot matching}
\label{sec:task}

We first describe unimodal one-shot matching and then extend it to the multimodal case.
As an example, we consider one-shot speech classification, illustrated on the left in Figure~\ref{fig:unimodal+multimodal}(a).
The model is shown a support set $\mathcal{S}$, containing one isolated spoken word with a text label for each of the $L$ word classes.
From this 
set, the model must learn a classifier $C_{\mathcal{S}}$ that can make predictions on an unseen test query $\mathbf{x}^*_{a}$. 
approach 
is to simply compare the query with each item in the support set and then predict the label of the closest item, as illustrated on the right in Figure \ref{fig:unimodal+multimodal}(a).

Figure \ref{fig:unimodal+multimodal}(b) illustrates 
\textit{multimodal} 
one-shot speech-image matching.
Instead of labelled examples, the
multimodal support set $\mathcal{S}=\{(\boldsymbol{\mathrm{x}}_a^{(i)}, \boldsymbol{\mathrm{x}}_v^{(i)})\}_{i=1}^L$ consists of pairs, where each isolated spoken word 
$\boldsymbol{\mathrm{x}}_a^{(i)}$ 
has a corresponding image 
$\boldsymbol{\mathrm{x}}_v^{(i)}$.
One pair is given for each of the $L$ classes.
At test time, the model is presented with an unseen spoken query $\boldsymbol{\mathrm{x}}^*_a$ and asked to determine the matching image in a test (or matching) set  $\mathcal{M}_v=\{(\boldsymbol{\mathrm{x}}_v^{(i)})\}_{i=1}^N$ of unseen images, as illustrated on the left in Figure \ref{fig:unimodal+multimodal}(b).
Neither the query $\mathbf{x}^*_{a}$ nor the matching set items $\mathcal{M}_v$ occur exactly in the support set $\mathcal{S}$.
To perform this task, we need to use
$\mathcal{S}$ to construct a distance metric $D_\mathcal{S}(\boldsymbol{\mathrm{x}}_a, \boldsymbol{\mathrm{x}}_v)$ between audio queries and test images.

The approach we use (originally proposed in~\cite{eloff_multimodal_2019}) is to
reduce the task 
to two unimodal comparisons, as shown on the right in Figure~\ref{fig:unimodal+multimodal}(b). 
First, 
we compare
the query $\boldsymbol{\mathrm{x}}^*_a$ to each $\boldsymbol{\mathrm{x}}_a^{(i)}$ in $\mathcal{S}$ to find the query's closest spoken neighbour 
in the support set. 
This closest neighbour's paired image is then compared to
each image $\boldsymbol{\mathrm{x}}_v^{(i)}$ in the matching set $\mathcal{M}_v$. 
This closest matching-set image is then selected as the model's prediction. In the figure, this is the image of the rightmost \textit{eight}.

We can also extend one-shot learning to $K$-shot learning.
In  unimodal $L$-way $K$-shot classification, the support set $\mathcal{S}$ contains
$L$ classes and $K$ labelled examples per class.
In multimodal $L$-way $K$-shot matching,
$\mathcal{S}=\{(\boldsymbol{\mathrm{x}}_a^{(i)}, \boldsymbol{\mathrm{x}}_v^{(i)})\}_{i=1}^{L \times K}$ consists of $K$ 
speech-image pairs for each of the $L$ classes.

%% file: Feature_extraction_methods_for_multimodal_one-shot_learning/Feature_extraction_methods_for_multimodal_one-shot_learning.tex
\graphicspath{{Feature_extraction_methods_for_multimodal_one-shot_learning/Figures/}}

\section{Feature representations}
\label{sec:models} 

\begin{figure*}[htb]

\begin{minipage}[b]{1.0\textwidth}
    \centering
    \includegraphics[width=0.75\textwidth]{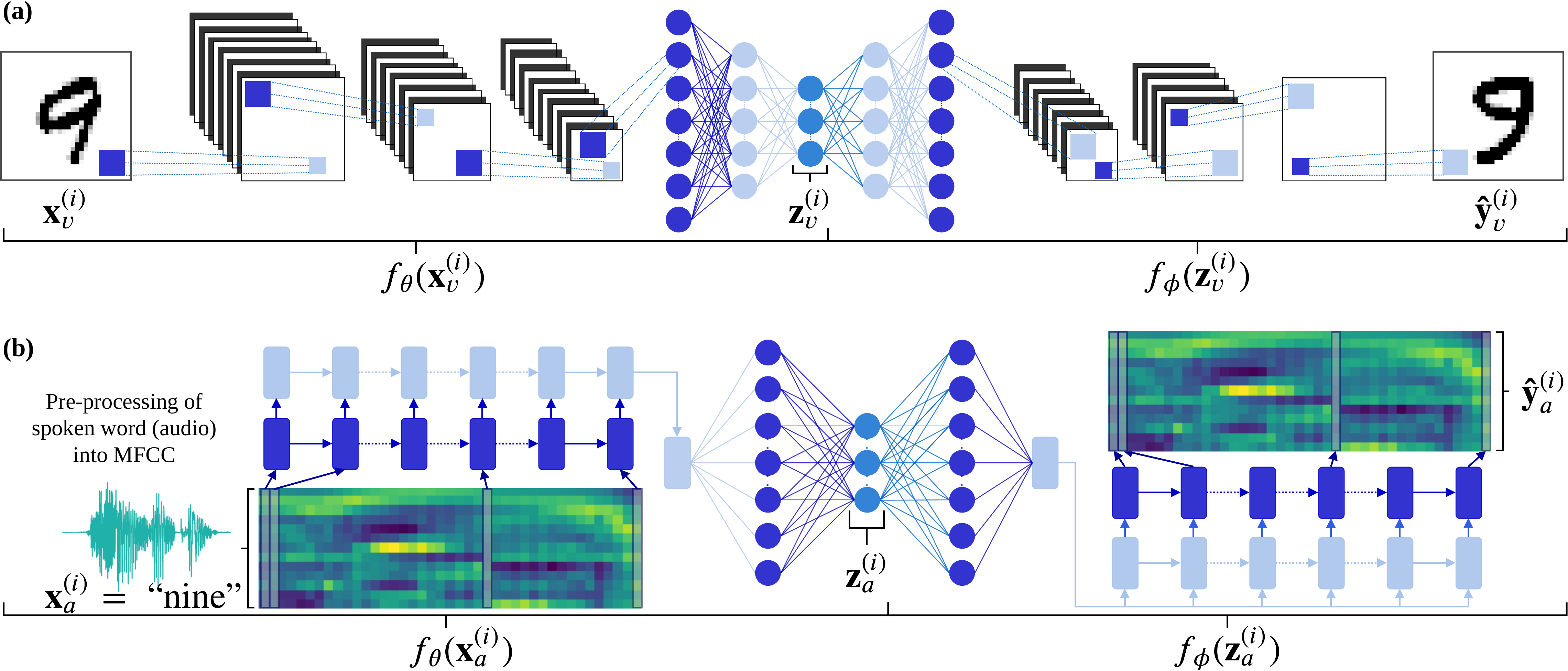}
        \vspace*{-6pt}
\end{minipage}

\caption{(a) Convolutional neural networks (CNNs) are used to learn 
feature representations for image data and (b) recurrent neural networks (RNNs) are used to learn feature representations for speech data.}
\label{fig:arch}

        \vspace*{-12pt}

\end{figure*}

In the 
description above we implicitly assume that we have a method or model that can measure similarity within a modality.
The aim of this paper is to consider 
different feature representations for these 
similarity comparisons, specifically comparing transfer learning (used in~\cite{eloff_multimodal_2019}) to unsupervised feature learning.
To compare the different features, we use the same framework as in~\cite{eloff_multimodal_2019} where multimodal one-shot learning is performed via two unimodal comparisons (as outlined above, Figure~\ref{fig:unimodal+multimodal}(b)-right).
Note that this is not an end-to-end approach; future work will explore learning direct cross-modal matching networks.

As a baseline, we use raw speech and image features directly (\S\ref{sec:raw}).
We then consider different neural networks to learn feature representations (\S\ref{sec:unsup_models} and \S\ref{sec:trans_models}).
We use separate networks for learning speech and image features.
For both the speech and vision models, we consider two settings: training on unlabelled in-domain data~(\S\ref{sec:unsup_models}) and training on labelled background data~(\S\ref{sec:trans_models}). 

\subsection{Raw feature matching}
\label{sec:raw}

As a nearest neighbour baseline, we use cosine distance over image pixels for image-to-image comparisons, and dynamic time warping (DTW) over MFCCs for speech-to-speech comparisons.

\subsection{Unsupervised models on unlabelled in-domain data}
\label{sec:unsup_models}

We consider two unsupervised models trained on unlabelled in-domain speech and vision data---data which includes unlabelled instances of  classes that we will see during one-shot testing.

An autoencoder (AE) 
is an unsupervised neural network which 
aims to reconstruct 
its input through a lower dimensional latent 
representation that acts as an information bottleneck \cite{chicco_deep_2014}.
As shown in Figure \ref{fig:arch}, the AE's encoder 
$f_\theta(\boldsymbol{\mathrm{x}}^{(i)})$ 
encodes the input $\boldsymbol{\mathrm{x}}^{(i)}$ to the feature representation $\boldsymbol{\mathrm{z}}^{(i)}$.
The decoder $f_\phi(\boldsymbol{\mathrm{z}}^{(i)})$ decodes $\boldsymbol{\mathrm{z}}^{(i)}$ to produce the output $\hat{\boldsymbol{\mathrm{y}}}^{(i)}$.
We use a 
squared loss
between the network's output $\hat{\boldsymbol{\mathrm{y}}}^{(i)}$ and the desired output $\boldsymbol{\mathrm{y}}^{(i)}$, 
i.e., $\ell = ||\boldsymbol{\mathrm{y}}^{(i)} - \hat{\boldsymbol{\mathrm{y}}}^{(i)}||^2_2$,
with the target set to $\boldsymbol{\mathrm{y}}^{(i)} = \boldsymbol{\mathrm{x}}^{(i)}$. 

The correspondence autoencoder (CAE) is identical to the AE but instead of reproducing the input $\boldsymbol{\mathrm{x}}^{(i)}$, it aims to reproduce another instance $\boldsymbol{\mathrm{x}}^{(i)}_{\textrm{pair}}$ of the same class as the input~\cite{kamper_unsupervised_2015}, i.e.\ we set the target $\boldsymbol{\mathrm{y}}^{(i)} = \boldsymbol{\mathrm{x}}^{(i)}_{\textrm{pair}}$ in the loss $\ell$.
The intuition 
is that the CAE will produce
features 
that are
invariant to properties not common to two inputs
while capturing aspects that are, such as the class.
We consider two variants of the CAE: one trained from scratch and another {pretrained} 
as an AE before switching to the CAE loss (denoted as AE-CAE).
To train the CAE, we need pairs of items of the same class. Since our in-domain data is unlabelled, we use cosine distance over pixels
to find image pairs that are most {alike}, 
and DTW to find 
spoken word pairs predicted to be of the same type.
{Speaker information is used to ensure that speech pairs are from different speakers.}

Using unlabelled in-domain image data, we train unsupervised vision networks with the AE, CAE and AE-CAE losses; we use the architecture shown in Figure~\ref{fig:arch}(a), with a convolutional neural network (CNN) encoder producing the latent feature vector, and a decoder with transposed convolutions.
Similarly, we use unlabelled in-domain speech data to train unsupervised speech networks using the AE, CAE and AE-CAE losses; we use an encoder recurrent neural network (RNN) producing the latent feature vector which is then used to condition a decoder RNN, as shown in Figure~\ref{fig:arch}(b).
These 
speech RNNs are similar to the acoustic embedding models of~\cite{chung_unsupervised_2016,wang_segmental_2018,holzenberger_learning_2018,kamper_truly_2019},
since they give a fixed-sized embedding for variable duration input.

\subsection{Transfer learning from labelled background data}
\label{sec:trans_models}

We next consider training 
supervised models on labelled background data. 
These 
datasets do not contain any instances of the target
one-shot classes. 
The idea is that features learned by such models would still be useful for determining similarity on unseen classes~\cite{vinyals_matching_2016}.
This is a form of \textit{transfer learning}~\cite{pan_survey_2009,donahue_decaf_2014}.

We specifically consider supervised classifier and Siamese neural networks, as in~\cite{eloff_multimodal_2019}.
We use identical architectures to
the encoder parts
of the networks in Figure \ref{fig:arch}. 
For the classifiers, we add a softmax 
layer after the feature embedding layer~$\boldsymbol{\mathrm{z}}^{(i)}$ and train the networks with the multiclass log loss.

A Siamese network does not classify an input, but measures similarity between inputs \cite{bromley_signature_1994}. 
The network consists of 
identical sub networks with shared parameters; each network maps its 
input to an embedding. 
Ideally, 
inputs of the same class should have similar 
embeddings and inputs of different classes should have different 
embeddings.
Say we have inputs $\boldsymbol{\mathrm{x}}$, $\boldsymbol{\mathrm{x}}_{\textrm{pair}}$ and $\boldsymbol{\mathrm{x}}_{\textrm{neg}}$, where $\boldsymbol{\mathrm{x}}$ and $\boldsymbol{\mathrm{x}}_{\textrm{pair}}$ are from the same class and $\boldsymbol{\mathrm{x}}$ and $\boldsymbol{\mathrm{x}}_{\textrm{neg}}$ are from different classes. 
We want the distance between the embeddings of $\boldsymbol{\mathrm{x}}$ and $\boldsymbol{\mathrm{x}}_{\textrm{pair}}$ to be smaller than those of $\boldsymbol{\mathrm{x}}$ and $\boldsymbol{\mathrm{x}}_{\textrm{neg}}$. 
We use the 
triplet
hinge 
loss $l(\boldsymbol{\mathrm{x}}, \boldsymbol{\mathrm{x}}_{\textrm{pair}}, \boldsymbol{\mathrm{x}}_{\textrm{neg}}) = \mathrm{max}\{0, m+d(\boldsymbol{\mathrm{x}}, \boldsymbol{\mathrm{x}}_{\textrm{pair}})-d(\boldsymbol{\mathrm{x}}, \boldsymbol{\mathrm{x}}_{\textrm{neg}})\} $, 
where $d(\boldsymbol{\mathrm{x}}_{1}, \boldsymbol{\mathrm{x}}_{2}) = \begin{Vmatrix}\boldsymbol{\mathrm{z}}_{1} - \boldsymbol{\mathrm{z}}_{2}\end{Vmatrix}_2^2$ is the squared Euclidean distance between the embeddings $\mathbf{z}_1$ and $\mathbf{z}_2$ of $\mathbf{x}_1$ and $\mathbf{x}_2$, respectively, and $m$ is a 
margin parameter~\cite{wang_learning_2014,hermann_multilingual_2014}.
To sample negative items, we use the online semi-hard mining scheme, where for each positive pair $(\mathbf{x}, \mathbf{x}_{\textrm{pair}})$, the most difficult negative pair $(\mathbf{x}, \mathbf{x}_{\textrm{neg}})$ is sampled (meeting some constraints)~\cite{schroff_facenet:_2015,hoffer_deep_2015,hermans_defense_2017}.

Again, separate classifier and Siamese vision CNNs and speech RNNs are trained on labelled background data.
We also consider supervised variants of the CAE and AE-CAE approaches, where instead of finding input-output training pairs based on their nearest neighbours (\S\ref{sec:unsup_models}), we train on ground truth pairs from the background data (these were not considered in~\cite{eloff_multimodal_2019}).
For all of the  models, 
we use the embedding $\boldsymbol{\mathrm{z}}^{(i)}$ as representation for unseen input~$\boldsymbol{\mathrm{x}}^{(i)}$.

%% file: Experiments/Experiments.tex
\graphicspath{{Experiments/Figures/}}

\section{Experimental setup}
\label{sec:setup}

\subsection{Data}

We follow the same 
setup as \cite{eloff_multimodal_2019}, using a dataset of
paired isolated spoken digits and handwritten digit images~\cite{kashyap_learning_2017}.
Speech data are parametrised as Mel-frequency cepstral coefficients (MFCCs). 
Image pixels are normalised to $[\textrm{0}, \textrm{1}]$.
We use the TIDigits corpus as our in-domain speech data; the corpus consists of spoken digit sequences from 326 speakers~\cite{leonard_r._gary_tidigits_1993}.
We split these sequences into isolated digits using forced alignments.
As our in-domain image data,
we use the MNIST corpus which contains $\textrm{28} \times \textrm{28}$ grayscale handwritten digit images~\cite{lecun_yann_gradient-based_1998}.
Although the TIDigits and MNIST datasets are labelled, note that we use it as unlabelled in-domain data for the models in \S\ref{sec:unsup_models}; we specifically train these unsupervised models on unlabelled isolated examples from the training subsets of these datasets. All 
one-shot evaluation experiments 
are then performed on 
the MNIST and TIDigits test~subsets.

For background speech data, we use the Buckeye corpus of English speech from 40 speakers \cite{pitt_mark_a._buckeye_2005}.
We use forced alignments to extract a set of labelled isolated words from this set.
For background image data, we use Omniglot~\cite{lake_human-level_2015}, containing 1623 types of handwritten characters which we invert and downsample to $\textrm{28} \times \textrm{28}$. 
We ensure that there are no instances of the target
digit classes in either the Buckeye or Omniglot background data.

\subsection{Models}

Neural networks  are implemented in TensorFlow and trained using 
Adam optimisation~\cite{kingma_adam_2015} with a learning rate of 10$^{-\textrm{3}}$.
Model hyperparameters were tuned using unimodal one-shot classification on test {subsets} of the background data, while early stopping was performed on validation subsets---neither of these 
background sets have item or class overlap with the final evaluation data.
We use a {feature} embedding dimensionality of 130 in all models to make results comparable. 
All speech RNNs take static MFCCs as input, but first and second order derivatives are used in the DTW baseline where it is
beneficial.

Unsupervised speech RNNs are trained on unlabelled isolated digits from the TIDigits training set using the AE, CAE and AE-CAE losses (\S\ref{sec:unsup_models}).
In all cases, the encoder and decoder each consists of three 400-unit RNN layers.
Unsupervised vision CNNs are trained with the AE, CAE and AE-CAE losses on unlabelled images from the MNIST training set.
The encoder consists of three convolutional layers with 3$\times$3 kernels and 32, 64 and 128 units; the decoder has the inverse architecture.

For transfer learning (\S\ref{sec:trans_models}), we train supervised classifier and Siamese  speech RNNs on labelled isolated words from the Buckeye training set. Similarly, we train supervised classifier and Siamese vision CNNs on Omniglot.
All these supervised models share the same structure as the encoder components from their unsupervised counterparts.
We also
train supervised
variants of the
CAE and AE-CAE speech and vision models
on the labelled background~data.

\subsection{Evaluation}
\label{sec:exp_eval}

We evaluate models averaged over 400 ``episodes''~\cite{vinyals_matching_2016}. 
To construct the support set, each multimodal episode randomly samples a 
spoken digit and paired image for each of the $L = \textrm{11}$ classes (``one" to ``nine", as well as ``zero" and ``oh").
A matching set is then sampled for testing, containing ten digit images
not in the support set. Finally, a spoken query is sampled, also
not in the support set.
The speech query then needs to be matched to the correct image in the matching set.
The matching set only contains ten digit images since there are only ten unique handwritten digit classes (both ``zero'' and ``oh'' are counted as correct if the image is that of a $0$). 
Within an episode, ten different query instances are also sampled while keeping the support and matching sets fixed.
We report unimodal 
and multimodal one- and five-shot matching accuracies with 95\% confidence intervals
averaged over 
five 
models trained with different seeds.

\begin{table}[!t]
    \mytable
    \caption{Unimodal one- and five-shot speech classification.} 
    \vspace*{-7.5pt}
    \eightpt
    \begin{tabularx}{1.0\linewidth}{@{}Llcc@{}}
        \toprule
        \multicolumn{2}{c}{\multirow{2}{*}{Model}} & \multicolumn{2}{c}{11-way accuracy (\%)}\\
        &  &  one-shot & five-shot \\
        \midrule
        Baseline & DTW & 65.90 & 89.45\\
        \addlinespace
        \multirow{4}{7em}{Transfer learning models}
& Classifier RNN & \textbf{86.87 $\pm$ 0.83} & \textbf{95.40 $\pm$ 0.50} \\
& Siamese RNN & 83.52 $\pm$ 2.56 & 94.34 $\pm$ 0.86 \\
& CAE RNN & 79.89 $\pm$ 1.32 & 92.16 $\pm$ 0.90 \\
& AE-CAE RNN & 80.02 $\pm$ 1.04 & 93.91 $\pm$ 0.25 \\
        \addlinespace
        \multirow{3}{7em}{Unsuper\-vised models}
& AE RNN & 53.82 $\pm$ 1.70 & 75.58 $\pm$ 1.54 \\
& CAE RNN & 75.80 $\pm$ 1.76 & 95.14 $\pm$ 0.80 \\
& AE-CAE RNN & 77.01 $\pm$ 1.29 & 93.30 $\pm$ 0.56 \\
        \bottomrule
    \end{tabularx}
    \label{tbl:speech}

    \vspace*{-6pt}
\end{table}

\section{Experimental Results}
\label{sec:experiments} 

\subsection{\textbf{\it K}-shot unimodal speech and image classification}
\label{sec:k-shot_speech}

We first consider unimodal results in isolation.
Table \ref{tbl:speech}
shows 
one- and five-shot speech classification results. 
All models except the AE RNN 
outperform the baseline. 
The classifier RNN achieves the highest 
accuracies, followed by the Siamese RNN.
In all cases, 
transfer learning models outperform their unsupervised counterparts, except for the five-shot CAE RNN.

For unimodal
image classification (not shown here), the trends are very similar, with the classifier and Siamese CNNs achieving accuracies of around 64\% and 84\% for the one- and five-shot cases, respectively. Again, these transfer learning models outperform all the unimodal unsupervised image models.

\subsection{\textbf{\it K}-shot multimodal speech and image matching}
\label{sec:k-shot_speech_image}

Table \ref{tbl:multimodal} shows multimodal one- and five-shot results.\footnote{Note that
the
results here are not directly comparable to that of~\cite{eloff_multimodal_2019}.
We found a small bug in the validation setup of~\cite{eloff_multimodal_2019}; the scores across models in~\cite{eloff_multimodal_2019} are comparable,
but lower scores are achieved when using the proper validation setup used in this paper.
We reran the code of~\cite{eloff_multimodal_2019} to confirm the scores reported here. 
}
In each case, the same model type is used to obtain speech and image features, e.g.\ the \textit{Classifier} row uses a {CNN} vision classifier to get image features with an {RNN} speech classifier for speech features.
In both one- and five-shot multimodal matching, the classifier performs best followed closely by the Siamese model.
None of the unsupervised models perform as well as these models obtained using transfer learning.
For the 
CAE and AE-CAE losses, the models trained using labelled background data also outperform the unsupervised variants.

\begin{table}[!t]
    \mytable
    \caption{Multimodal one- and five-shot speech-image matching.} 
    \vspace*{-7.5pt}
    \eightpt
    \begin{tabularx}{1.0\linewidth}{@{}Llcc@{}}
        \toprule
        \multicolumn{2}{c}{\multirow{2}{*}{Model}} & \multicolumn{2}{c}{11-way accuracy (\%)}\\
        &  &  one-shot & five-shot \\
        \midrule
        Baseline & DTW + Pixels & 31.80 & 41.88\\
        \addlinespace
        \multirow{4}{7em}{Transfer learning models}
& Classifier~\cite{eloff_multimodal_2019} & \textbf{56.80 $\pm$ 1.19} & \textbf{59.67 $\pm$ 1.73} \\
& Siamese~\cite{eloff_multimodal_2019} & 54.83 $\pm$ 1.80 & 59.25 $\pm$ 0.79 \\
& CAE & 46.60 $\pm$ 0.69 & 53.82 $\pm$ 1.07 \\
& AE-CAE & 48.15 $\pm$ 1.21 & 56.81 $\pm$ 1.21 \\
        \addlinespace
        \multirow{3}{7em}{Unsuper\-vised models}
& AE & 28.99 $\pm$ 0.84 & 38.68 $\pm$ 1.51 \\
& CAE & 42.75 $\pm$ 0.62 & 52.15 $\pm$ 0.69 \\
& AE-CAE & 42.81 $\pm$ 1.01 & 50.28 $\pm$ 0.29 \\
        \bottomrule
    \end{tabularx}
    \label{tbl:multimodal}
\end{table}

\begin{table}[!t]
    \mytable
    \caption{Multimodal one- and five-shot speech-image matching using models that combine transfer and unsupervised learning.
} 
    \vspace*{-7.5pt}
    \eightpt
    \begin{tabularx}{1.0\linewidth}{@{}Lcc@{}}
        \toprule
        \multicolumn{1}{c}{\multirow{2}{*}{Model}} & \multicolumn{2}{c}{11-way accuracy (\%)}\\
        &  one-shot & five-shot \\
        \midrule
        Baseline: DTW + Pixels & 31.80 & 41.88\\
        \addlinespace
        Transfer learning: Classifier~\cite{eloff_multimodal_2019}  & \textbf{56.80 $\pm$ 1.19} & \textbf{59.67 $\pm$ 1.73} \\
        \addlinespace
        CAE with cosine pairs & 42.75 $\pm$ 0.62 & 52.15 $\pm$ 0.69 \\
        CAE with classifier pairs & 48.66 $\pm$ 1.14 & 55.59 $\pm$ 0.71 \\
        Transfer learning + CAE fine-tuning & 54.32 $\pm$ 2.19 & 59.37 $\pm$ 1.80 \\
        \addlinespace
        CAE with oracle pairs & 89.19 $\pm$ 0.69 & 92.81 $\pm$ 0.47 \\
        \bottomrule
    \end{tabularx}
    \label{tbl:combined}

    \vspace*{-6pt}
\end{table}

\subsection{Towards combined transfer and unsupervised learning}

It is evident that the transfer learning approach originally followed in~\cite{eloff_multimodal_2019} outperforms the unsupervised approach developed here.
However, the two methodologies might be complementary: transfer learning from background data could capture general properties within a particular modality, while unsupervised learning on unlabelled in-domain data could provide a way to tailor representations to a specific test setting.

As an initial investigation,
we propose two combined models here, with results given in Table~\ref{tbl:combined}.
The \textit{CAE with cosine pairs} (row 3) is repeated from Table~\ref{tbl:multimodal}.
Instead of finding nearest neighbours using cosine distance, we use representations from the classifier (trained on background data) to find pairs in the unlabelled in-domain data for training a CAE (as with the standard CAE, speaker information is still used to ensure that pairs are from different speakers).
We see that this \textit{CAE with classifier pairs} (row 4) gives a small improvement over the standard CAE.
By additionally initialising the CAE by training it on the labelled background data 
and then fine-tuning it on the in-domain data, 
we get a further improvement (\textit{Transfer learning + CAE fine-tuning}, row 5).
Neither of these approaches, however, outperform the transfer learned classifier (row 2).

In order to see if it is at all possible to achieve better performance with the CAE by using more accurate training pairs, we also give the performance of a CAE trained only using correct pairs
in the last row of Table~\ref{tbl:combined}.
We see that this oracle model outperforms all other approaches, indicating that, if we were able to improve the 
CAE's training pairs, we might be able to take advantage of an unsupervised learning scheme.

%% file: Conclusion/Conclusion.tex
\section{Conclusion}
\label{sec:conclusion}

We have compared existing and new models for few-shot multimodal speech-image matching.
Transfer learning from background data consistently outperformed unsupervised modelling on unlabelled in-domain data on a multimodal one-shot matching benchmark.
We also proposed two approaches for combining transfer and unsupervised learning.
Although neither improved the best transfer learning approach, performance improved over the standard unsupervised approach.
We will therefore also consider other approaches for combining the methodologies in future work.
Building on models which directly maps images and unlabelled speech into a joint space~\cite{ngiam_multimodal_2011,harwath_unsupervised_2016,leidal_learning_2017,harwath_jointly_2018}, we will also consider end-to-end solutions for multimodal one-shot learning.\blfootnote{This work is supported in part by the National Research Foundation of South Africa (grant number: 120409), a Google Faculty Award for HK, a DST CSIR scholarship for LN, and funding from Saigen.}